\RequirePackage[svgnames,table]{xcolor}

\documentclass[11pt,letterpaper,logo]{yalearxiv}

\usepackage{graphicx}
\usepackage{float,epstopdf}
\usepackage{bbm}

\usepackage{microtype}

\usepackage{natbib}
\setcitestyle{square}

\usepackage{subcaption}
\usepackage{booktabs}
\usepackage{amsmath}
\usepackage{amssymb}
\usepackage{mathtools}
\usepackage{amsthm}
\usepackage{dsfont}
\usepackage{multicol}
\usepackage{makecell}
\usepackage{multirow}
\usepackage{amsfonts}
\usepackage{mathrsfs}
\usepackage[amssymb, thickqspace]{SIunits}
\usepackage{enumitem}
\usepackage{pgfplotstable}
\usepackage{lipsum}		

\usepackage{microtype}
\usepackage{graphicx}
\usepackage{booktabs} 
\usepackage[table]{xcolor}
\usepackage{arydshln}
\usepackage[normalem]{ulem} 

\usepackage{cases}
\usepackage{wrapfig}

\usepackage{url}

\usepackage{thmtools}
\usepackage{thm-restate}
\usepackage{tabu}

\definecolor{huskypurple}{HTML}{4B2E83}

\usepackage{titletoc}
\usepackage{listings}
\lstdefinestyle{promptstyle}{
  basicstyle=\ttfamily\footnotesize,
  breaklines=true,
  breakautoindent=false,
  breakindent=0pt,
  postbreak=\mbox{\textcolor{gray}{$\hookrightarrow$}\space},
  columns=fullflexible,
  keepspaces=true,
  frame=single,
  framesep=5pt,
  xleftmargin=6pt,
  xrightmargin=6pt,
  aboveskip=8pt,
  belowskip=8pt,
  showstringspaces=false,
}
\makeatletter
\def\munderbar#1{\underline{\sbox\tw@{$#1$}\dp\tw@\z@\box\tw@}}
\makeatother

\AddToHook{cmd/appendix/before}{%
  \setcounter{axiom}{0}%
}

\newcommand{\be}{\begin{equation}}
\newcommand{\ee}{\end{equation}}
\newcommand{\bea}{\begin{equation*}\begin{aligned}}
\newcommand{\eea}{\end{aligned}\end{equation*}}

\newtcolorbox{simpleElegantQuote}{
    colback=AliceBlue!50!White,
    colframe=RoyalBlue!75!Black,
    boxrule=0.5pt,
    arc=2mm,
    boxsep=4pt,
    left=10pt, right=10pt,
    top=8pt, bottom=8pt,
    fontupper=\itshape,
}

\title{openJiuwen: Beyond Static Harnesses for Long-Horizon Coding Agents}

\runningtitle{openJiuwen: Beyond Static Harnesses for Long-Horizon Coding Agents}


\usepackage{fontawesome5}   
\usepackage{float}

\definecolor{yaleblue}{RGB}{0,58,112}

\author{openJiuwen Team\\
Huawei Technologies Co., Ltd.\\
\faGithub~\textbf{Source Code:} \href{https://github.com/openJiuwen-ai/jiuwenswarm}{\texttt{https://github.com/openJiuwen-ai/jiuwenswarm}}
}

\hypersetup{colorlinks=true, linkcolor=blue!50!black, citecolor=blue!50!black,
            urlcolor=blue!50!black}

\begin{document}

\begin{abstract}
\vspace{-1mm}
\begin{center}
    \section*{Abstract}
\end{center}

Long-horizon coding agents operate over evolving repository states while increasingly relying on heterogeneous capabilities, delegated agents, and multi-agent coordination. These trends pose two complementary challenges for the agent harness. First, developers need to compose capabilities, reconfigure execution logic, and scale increasingly complex agent systems without repeatedly rebuilding orchestration. Second, complex coding tasks continuously produce new evidence---such as semantic diagnostics, execution outcomes, task progress, and changing context relevance---that should dynamically influence subsequent runtime decisions. We characterize these challenges as \textbf{Structural Composability} and \textbf{Runtime Adaptivity}. We present \textbf{openJiuwen}, an \textbf{open-source harness} designed for both developer composability and adaptive task execution. openJiuwen provides a shared execution substrate and Rail-based capability composition across single agents, delegated sub-agents, and Swarm Flow, enabling developers to construct sophisticated agent harnesses under common execution semantics. It further adapts framework-controlled runtime decisions around a fixed model policy, allowing evolving evidence to dynamically affect context, feedback, and task control toward successful completion. We systematically evaluate openJiuwen on \textbf{SWE-bench Verified} and \textbf{Terminal-Bench 2.1}, where it achieves \textbf{82.6\%} and \textbf{87.19\%}, respectively, exceeding the strongest selected official-leaderboard point estimates by \textbf{3.4} and \textbf{3.39 percentage points}. These results show that openJiuwen achieves strong performance on complex coding tasks while providing a composable and adaptive harness design.

\end{abstract}

\maketitle

\section{Introduction}
\label{sec:intro}

Large language models (LLMs) are transforming automated software engineering from single-turn code generation into \emph{long-horizon} coding agents that interact with repositories, tools, and execution environments over extended trajectories. Complex coding issues require more than locally correct actions: an agent must maintain coherent behavior while code state, diagnostics, task progress, and relevant context evolve throughout the run. As agents incorporate planning, memory, verification, context engineering, delegation, and multi-agent coordination, the \emph{agent harness} increasingly becomes a first-class systems layer that determines how these capabilities are assembled and how execution is controlled.

Recent agent harnesses already expose increasingly sophisticated mechanisms, but they organize them through distinct architectural choices. \textbf{Pi}\footnote{\url{https://github.com/earendil-works/pi}} emphasizes a minimal, provider-flexible coding runtime whose advanced behaviors are introduced via extensions. \textbf{DeepSeek Harness}\footnote{\url{https://github.com/deepseek-ai/deepseek-harness}} pushes further toward plugin-level composability, making core components—such as model adapters, tools, session states, and the execution loop—fully replaceable. \textbf{DeerFlow}\footnote{\url{https://github.com/bytedance/deer-flow}} focuses on long-horizon hierarchical orchestration through a lead agent and delegated sub-agents, while \textbf{Codex}\footnote{\url{https://github.com/openai/codex}} and \textbf{Claude Code}\footnote{\url{https://github.com/anthropics/claude-code}} integrate delegation, context management, skills/hooks, granular permissions, and lifecycle controls into mature developer-facing coding systems. Together, these systems demonstrate that the harness is no longer merely a thin tool wrapper around an LLM. As agent harnesses accumulate increasingly rich features, the harness itself becomes a new systems problem: how to organize, compose, and adapt these mechanisms coherently throughout long-horizon execution.

\begin{figure}[!t]
    \centering
    \includegraphics[width=\linewidth]{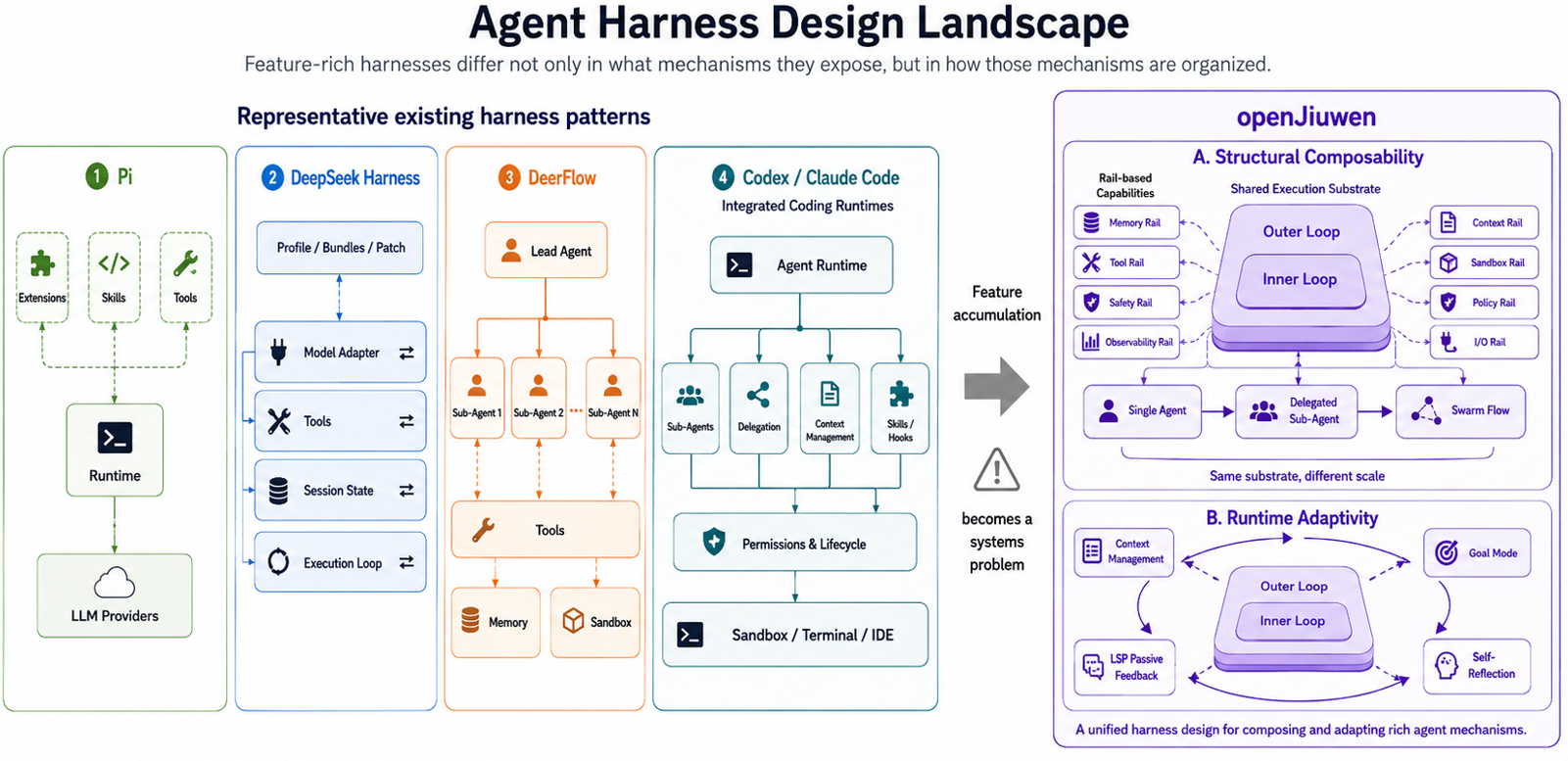}
    \caption{Architectural patterns of representative agent harnesses and the design of openJiuwen. Existing harnesses expose increasingly rich mechanisms through different architectural choices, while openJiuwen organizes harness complexity around Structural Composability and Runtime Adaptivity.}
    \label{fig:harness-comparison}
\end{figure}

The first challenge is therefore \textbf{developer complexity}. As agent harnesses accumulate increasingly rich features, extensibility alone does not ensure that these capabilities remain easy to organize and combine. Plugins, hooks, sub-agents, and workflow operators may each be individually extensible, yet integrating them across different coordination patterns can introduce fragmented execution paths and bespoke orchestration logic. The key question is therefore whether capabilities and agent topologies can be composed over a common execution model rather than requiring separate execution semantics for each configuration. We refer to this property as \textbf{Structural Composability}: the ability to assemble and reconfigure capabilities and execution units over a shared execution substrate, from a single agent to delegated sub-agents and programmable multi-agent flows. Its goal is developer-facing: to let increasingly sophisticated agent systems be constructed and evolved without coupling each new configuration to a separate execution architecture.

The second challenge is \textbf{execution uncertainty}. For complex coding issues, critical evidence is revealed only while the task is being solved, including semantic diagnostics, test outcomes, intermediate results, progress signals, and changes in context relevance or resource pressure. Decisions made before execution therefore become progressively stale. A harness must use newly available evidence to adapt framework-controlled decisions such as context construction, semantic feedback, task continuation, and stopping. We refer to this capability as \textbf{Runtime Adaptivity}. Its goal is task-facing: to steer the execution trajectory toward successful task completion by adapting runtime decisions around a fixed model policy as the task state evolves.

To address these challenges, we present \textbf{openJiuwen}, an \textbf{open-source harness} organized explicitly around Structural Composability and Runtime Adaptivity, as illustrated in Figure~\ref{fig:harness-comparison}. For \textbf{Structural Composability}, openJiuwen provides a shared Inner Loop/Outer Loop execution substrate together with the Rail mechanism for attaching capabilities through lifecycle hooks, ordered composition, and visibility gating. The same execution semantics are reused across single agents, delegated sub-agents, and Swarm Flow, allowing developers to scale agent structure without introducing a separate execution engine at each level. For \textbf{Runtime Adaptivity}, openJiuwen changes framework-controlled runtime state rather than model parameters: Context Management adapts the information exposed to the model, Goal Mode controls task acceptance and stopping, LSP-Driven Passive Feedback injects filtered semantic diagnostics into subsequent execution, and Self-Reflection distills completed trajectories into reusable experience for later tasks. Together, these mechanisms provide a unified harness architecture for composing complex agent structures and adapting their execution as runtime evidence emerges.

Our contributions are summarized as follows:

\begin{itemize}
    \item \textbf{Developer-facing structural composability.} We introduce a shared execution architecture that combines nested Inner Loop/Outer Loop control with Rail-based capability composition and reuses the same execution semantics across single agents, delegated sub-agents, and Swarm Flow. This design reduces the need for topology-specific execution engines and lowers the integration and orchestration burden of constructing and evolving sophisticated agent harnesses.

    \item \textbf{Task-oriented runtime adaptivity.} We formulate runtime adaptation as changes to framework-controlled execution state around a fixed model policy and instantiate it through Context Management, Goal Mode, LSP-Driven Passive Feedback, and Self-Reflection. These mechanisms allow evidence emerging during complex coding trajectories to influence subsequent execution through context construction, semantic feedback, continuation and stopping, and cross-task experience reuse.

    \item \textbf{Evaluation on complex agent tasks.} We systematically evaluate openJiuwen on \textbf{SWE-bench Verified} and \textbf{Terminal-Bench 2.1}, where it achieves \textbf{82.6\%} and \textbf{87.19\%}, respectively, exceeding the strongest selected official-leaderboard results by \textbf{3.4} and \textbf{3.39 percentage points}. Further analysis on SWE-bench Verified shows that openJiuwen remains competitive as task duration increases, supporting its effectiveness on long-horizon coding tasks.
\end{itemize}

\section{Related Work}
\label{sec:related_work}

\subsection{Coding-Agent Architectures and Scaffolds}

Recent work has advanced repository-level coding agents through different scaffold designs. SWE-agent~\cite{yang2024sweagent} develops dedicated interfaces for repository interaction, while CodeAct and OpenHands~\cite{wang2024executable,wang2025openhands} provide executable action spaces and general-purpose development environments. Other systems strengthen specific software-engineering capabilities, including program-structure-aware localization in AutoCodeRover~\cite{zhang2024autocoderover}, repository exploration in LingmaAgent~\cite{ma2025lingmaagent}, autonomous repair planning in RepairAgent~\cite{bouzenia2025repairagent}, fixed repair pipelines in Agentless~\cite{xia2025agentless}, and test-time generation and selection in Trae Agent~\cite{traeresearchteam2025traeagent}. Beyond single-agent scaffolds, Swarm Skills~\cite{zhang2026swarmskills} represents multi-agent roles, workflows, and coordination constraints as portable, self-evolving specifications. 
These approaches primarily improve particular interfaces, capabilities, or workflows; comparatively less attention has been paid to the harness as a reusable execution substrate for composing heterogeneous capabilities and scaling shared execution semantics from single agents to delegated and multi-agent execution.

\subsection{Runtime Behavior for Long-Horizon Coding Agents}

Long-horizon coding agents often revisit prior states or repeat work during execution. Empirical studies identify recurring action patterns and show that failed trajectories are typically longer and more variable even when relevant files have already been localized~\cite{lindenbauer2025complexitytrap,majgaonkar2025understanding}. Moving from diagnosis to intervention, Ledger~\cite{wang2026turning} maintains explicit execution state over observations, modifications, and prior actions, using it both to inform subsequent decisions and to mediate redundant commands. Collaborative Optimization~\cite{zhang2025collaborative} further optimizes agent workflows through joint adaptation of workflow structures, prompts, demonstrations, and routing decisions. These works highlight the importance of runtime control beyond isolated model reasoning, but primarily focus on particular execution-state or workflow-optimization mechanisms. Long-horizon coding additionally exposes evolving semantic feedback, task progress, and context relevance, motivating broader mechanisms that adapt execution as new information emerges. 

\subsection{Context Management for Long-Horizon Agents}

Another line of work manages long-horizon execution by controlling what information remains visible to the model. SWE-AGILE~\cite{lian2026sweagile} compresses older reasoning while retaining recent interactions, SWE-Pruner~\cite{wang2026swepruner} selects task-relevant context, and AgentDiet~\cite{xiao2026agentdiet} removes redundant or expired trajectory content; simple observation masking can also achieve competitive efficiency~\cite{bouzenia2025understanding}. Prompt-side adaptation provides another mechanism for changing model-visible inputs: P3~\cite{zhang2025p3} jointly optimizes system and user prompts and further performs query-dependent online prompting. These methods demonstrate the importance of adapting model inputs, but primarily operate on what is presented to the model. Long-horizon execution also produces evolving semantic feedback, task-progress signals, and reusable experience, motivating runtime adaptation beyond model-input adaptation alone.

\paragraph{Positioning of openJiuwen.}
openJiuwen complements these efforts by treating the harness itself as the primary systems layer. It provides a shared execution substrate for composing and scaling agent capabilities, while allowing runtime feedback, progress, and context changes to directly shape subsequent execution.

\section{Method: Composable Structure and Adaptive Runtime Control}
\label{sec:method}

openJiuwen is organized around two complementary dimensions. \emph{Structural composability} governs how capabilities and agent structures are composed over a shared execution substrate, from single agents to sub-agents and multi-agent systems. \emph{Runtime adaptivity} governs how execution changes as new information emerges, including context, semantic feedback, task progress, and reusable experience. The former defines how the system is composed; the latter adapts behavior over trajectories produced by that structure.

\begin{figure}[!t]
\centering
\includegraphics[width=\linewidth]{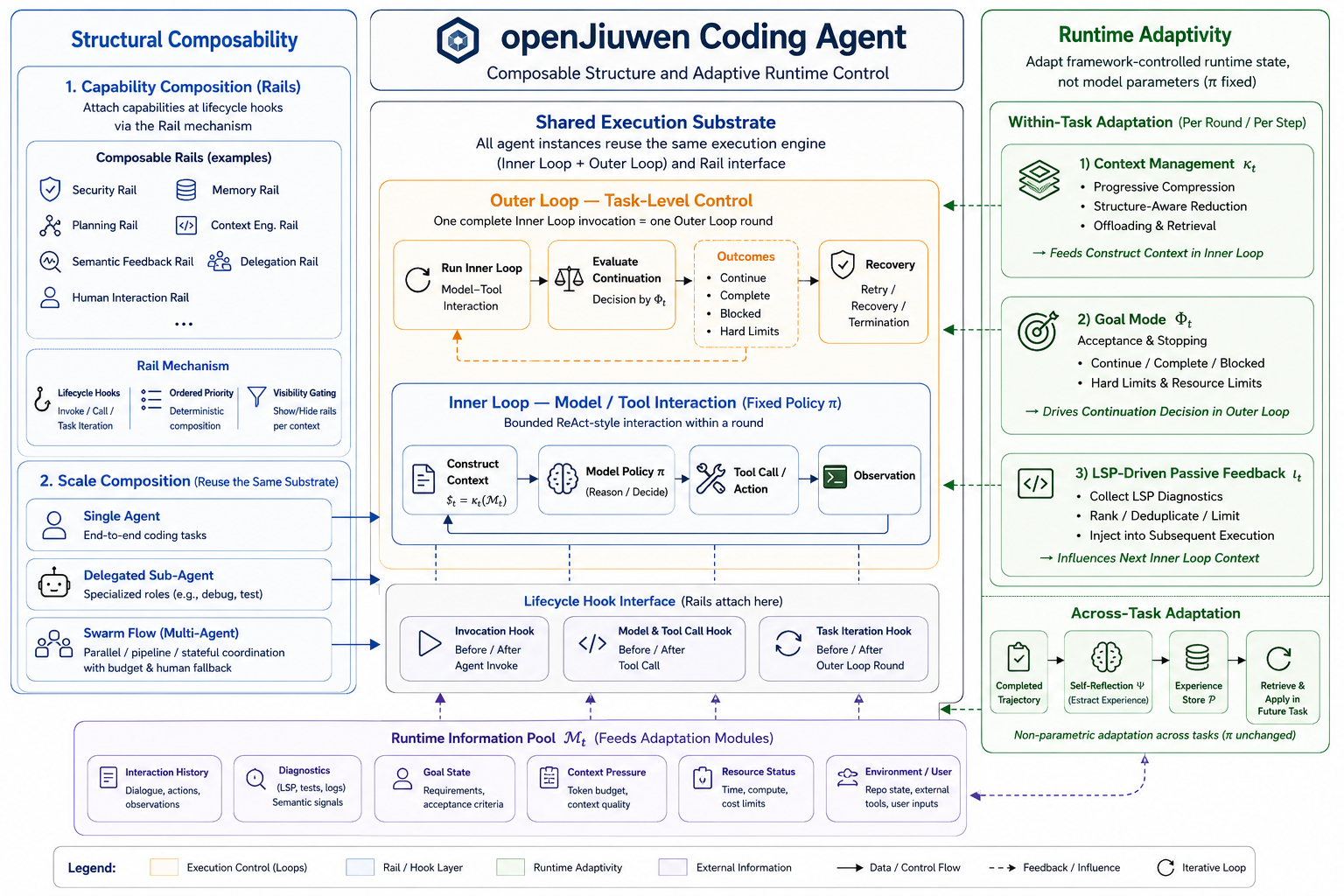}
\caption{
System overview of openJiuwen. Structural composability combines capability
composition through the Rail mechanism with scale reuse from single agents
and delegated sub-agents to Swarm Flow. All agent instances reuse the shared
Inner Loop/Outer Loop execution core, while runtime adaptivity is realized
through Context Management, Goal Mode, LSP-Driven Passive Feedback, and
Self-Reflection. The illustrated Rails and runtime mechanisms are
representative rather than exhaustive.
}
\label{fig:openjiuwen-overview}
\end{figure}

Figure~\ref{fig:openjiuwen-overview} provides a system-level view of both dimensions and their shared execution substrate.

\subsection{Structural Composability}
\label{sec:structural}

Every agent instance---whether standalone, delegated as a sub-agent, or invoked
within a Swarm Flow---reuses the same two primitives: the Inner Loop/Outer Loop
execution engine of \S\,\ref{sec:dualloop} and the Rail-based capability interface
of \S\,\ref{sec:rail}. Scaling from one agent to many therefore changes how agents and
capabilities are composed without introducing a separate agent execution
engine.

\subsubsection{Inner Loop and Outer Loop: Layered Execution}
\label{sec:dualloop}

openJiuwen separates agent execution into two nested loops with distinct
responsibilities. The Inner Loop handles model--tool interaction, while the
Outer Loop handles task-level continuation and execution control.

\paragraph{Inner Loop.}
The Inner Loop follows a bounded ReAct-style interaction. At each step, the
framework constructs the model-visible context, invokes policy $\pi$, executes
requested tool calls, records their observations, and exposes those
observations to subsequent model steps. The interaction continues until the
model returns an answer, execution is interrupted, or the configured step
limit is reached.

Lifecycle callbacks are triggered around model and tool boundaries, allowing
Rails to observe or modify execution without embedding capability-specific
logic in the loop itself.

\paragraph{Outer Loop.}
One complete Inner Loop invocation constitutes an Outer Loop round. After each
round, the framework evaluates whether execution should terminate or another
round should begin. The Outer Loop therefore handles task-level continuation
rather than step-level reasoning.

Continuation decisions are supplied through composable evaluators, including
semantic completion signals, resource limits, and user-defined conditions.
Goal Mode provides a stronger task-level completion mechanism and is described
in \S\,\ref{sec:goal}. The Outer Loop also provides a boundary for execution-level
control such as exceptions and bounded retry or recovery, separating
infrastructure failures from additional agent reasoning.

Input arriving during an active model interaction is incorporated at a later
execution boundary rather than mutating an in-flight call. The two loops thus
maintain separate control granularities: the Inner Loop governs local
reasoning and tool use, while the Outer Loop governs task-level continuation
and recovery.

\subsubsection{Rail: Ordered Capability Composition}
\label{sec:rail}

Cross-cutting capabilities such as security, memory, planning, context
engineering, semantic feedback, delegation, and human interaction require
visibility into execution without becoming part of the core execution
algorithm. openJiuwen therefore exposes lifecycle hooks and implements such
capabilities as Rails attached to these hooks.

Let
\begin{equation}
\label{eq:hook-space}
\mathcal{H} = \mathcal{H}_{\mathrm{invoke}} \cup \mathcal{H}_{\mathrm{call}} \cup \mathcal{H}_{\mathrm{task}}
\end{equation}
denote the lifecycle-hook space, corresponding respectively to
invocation-level, model-/tool-call-level, and task-iteration-level events.

A Rail is represented abstractly as
\begin{equation}
\label{eq:rail}
\rho = \left( \mathcal{H}_\rho, f_\rho, p_\rho \right),
\end{equation}
where $\mathcal{H}_\rho \subseteq \mathcal{H}$ specifies the lifecycle boundaries
to which the Rail attaches, $f_\rho$ is its handler, and $p_\rho$ is its
declared priority.

At each lifecycle hook, attached Rails execute according to their declared
priorities, with equal-priority cases resolved deterministically. When the
merge policy permits replacement, this ordering also provides controlled
overriding between capabilities. Composition order is therefore explicit
metadata rather than an accidental consequence of capability-specific branches
inside the execution loop.

Adding or removing a capability changes the Rail configuration while leaving
the Inner Loop/Outer Loop unchanged. Model- and tool-exception hooks similarly
provide uniform interception points through which Rails can participate in
failure handling while the framework retains control of execution.

\paragraph{Capability gating.}
Rail composition is complemented by visibility gating. Let
$g(\rho,u) \in \{0,1\}$ indicate whether Rail $\rho$ is visible to execution
subject $u$. The subject-specific capability configuration is
\begin{equation}
\label{eq:gated-rail}
\mathcal{R}(u) = \left\{ \rho \in \mathcal{R} \mid g(\rho,u) = 1 \right\}.
\end{equation}

The same principle applies to tools. Standalone agents, delegated sub-agents,
leaders, and other participating agents can therefore share the same execution
substrate while exposing different capabilities. Visibility gating supports
bounded recursive delegation, progressive capability disclosure, and
role-specific capability isolation without changing the underlying execution
engine.


\subsubsection{Swarm Flow: Composable Multi-Agent Coordination}
\label{sec:scale}

\begin{figure}[!t]
\centering
\includegraphics[width=\linewidth]{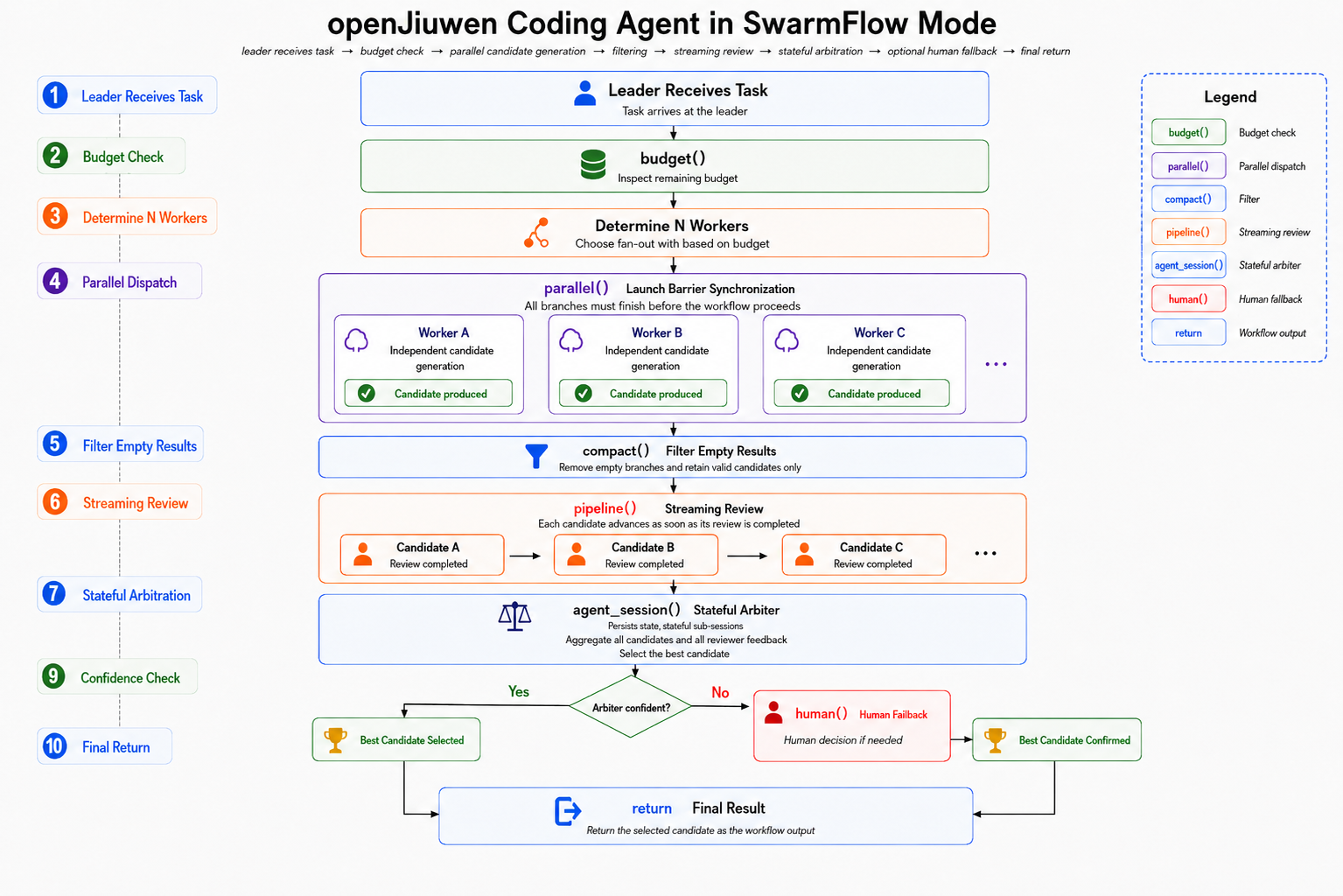}
\caption{
A Coding Agent implemented with openJiuwen Swarm Flow. The example illustrates
how the provided coordination operators can be composed into a
task-specific multi-agent workflow.
}
\label{fig:openjiuwen-swarmflow}
\end{figure}

\emph{Swarm Flow} extends the shared execution substrate to multi-agent systems
through a set of composable coordination operators. Rather than prescribing a
fixed multi-agent architecture, it allows developers to construct
task-specific coordination flows by combining operators for resource control,
parallel execution, result processing, stateful coordination, and fallback.

The core operators include:
\begin{itemize}
    \item \texttt{budget()} exposes the remaining execution budget to downstream control logic;
    \item \texttt{parallel()} dispatches multiple agent branches concurrently and synchronizes their completion;
    \item \texttt{compact()} filters invalid or empty branch results before downstream processing;
    \item \texttt{pipeline()} streams intermediate results through subsequent processing stages as they become available;
    \item \texttt{agent\_session()} maintains a stateful agent session across multiple stages of the flow;
    \item \texttt{human()} introduces optional human intervention when automated coordination is insufficient; 
    \item \texttt{return} terminates the flow and exposes the final result.
\end{itemize}

Figure~\ref{fig:openjiuwen-swarmflow} shows one concrete Coding Agent built
with these operators. The leader queries \texttt{budget()} and uses the
remaining budget to determine the number of workers. \texttt{parallel()}
launches these workers to generate independent candidates, while
\texttt{compact()} removes empty results and \texttt{pipeline()} streams
valid candidates through review. A stateful arbiter implemented with
\texttt{agent\_session()} aggregates candidate results and reviewer feedback.
If its confidence is insufficient, the flow can invoke \texttt{human()} before
producing the final \texttt{return}.

This Coding Agent is one realization of Swarm Flow. Developers can rearrange,
omit, and combine these operators to construct their own multi-agent
coordination strategies.

\subsection{Runtime Adaptivity}
\label{sec:adaptivity}

Structural composability determines how agents and capabilities are assembled,
whereas runtime adaptivity determines how their effective behavior changes as
execution unfolds. openJiuwen considers adaptation at two time scales. Within
a task, the runtime adjusts context construction, semantic feedback, and
acceptance/stopping decisions according to newly available execution evidence.
Across tasks, completed trajectories can be distilled into reusable experience
for later executions.

Both forms of adaptation operate around a fixed model policy $\pi$. openJiuwen
changes framework-controlled runtime state rather than model parameters.

\begin{figure}[!htb]
    \centering
    \includegraphics[width=0.85\linewidth]{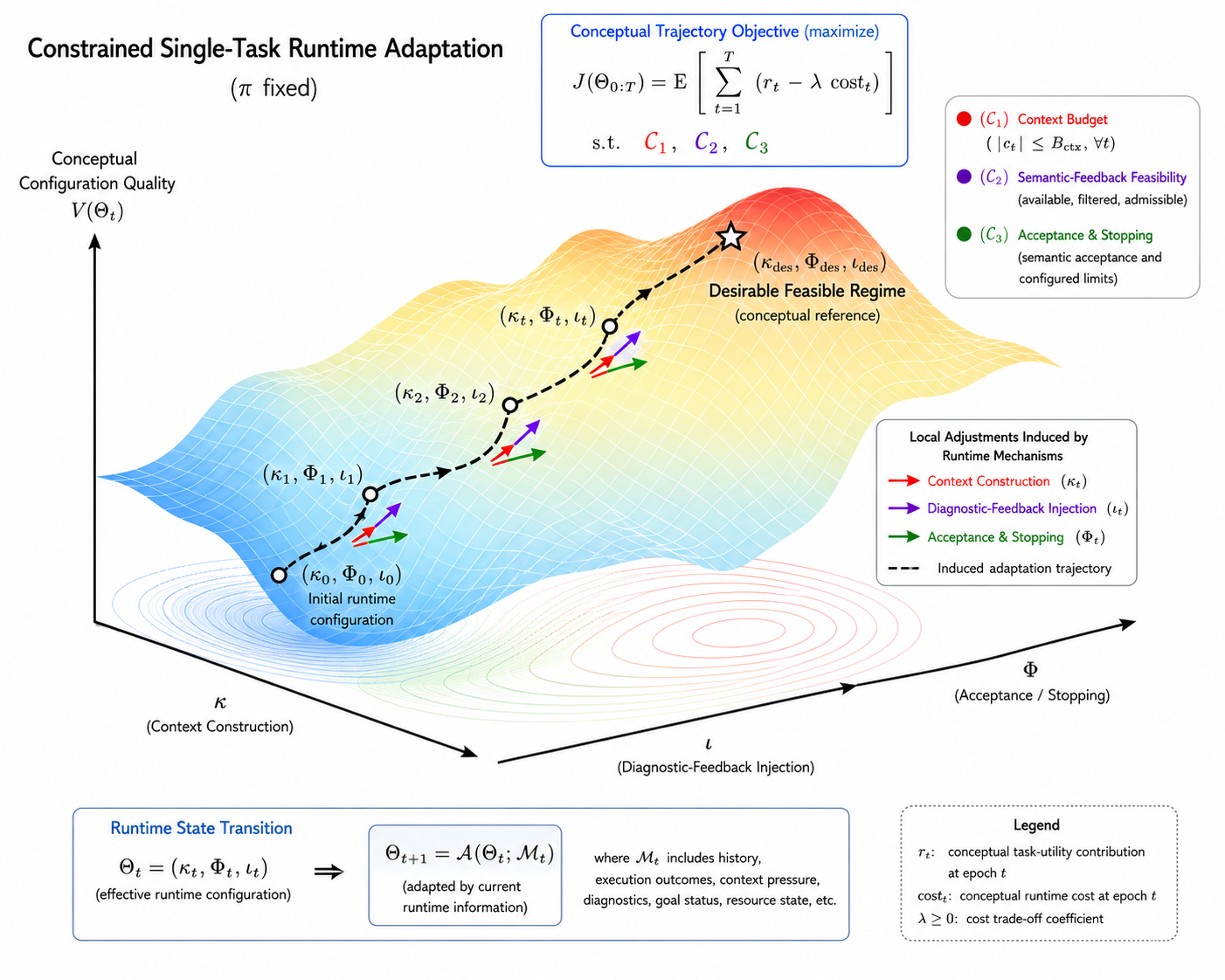}
    \caption{
    Conceptual view of constrained single-task runtime adaptation under a fixed
    model policy $\pi$. An effective runtime configuration
    $\Theta_t=(\kappa_t,\Phi_t,\iota_t)$ captures context construction,
    acceptance/stopping, and diagnostic-feedback injection. The surface is a
    schematic projection used to visualize relative configuration quality,
    while the trajectory-level objective $J(\Theta_{0:T})$ evaluates the
    execution trajectory as a whole. Runtime evidence induces state-dependent
    adaptation subject to context-budget, semantic-feedback, and stopping
    constraints.
    }
    \label{fig:runtime-adaptivity}
\end{figure}

\subsubsection{Constrained Online Runtime Adaptation}
\label{sec:runtime-opt}

We use constrained online optimization as a conceptual lens for single-task
runtime adaptivity rather than as a numerical optimization algorithm. At
runtime control epoch $t$, let $\mathcal{M}_t$ denote the information currently
available to the framework, including interaction history, diagnostics, goal
state, context pressure, and resource status. Context construction produces
\begin{equation}
\label{eq:context-constructor}
c_t = \kappa_t(\mathcal{M}_t),
\end{equation}
and the fixed model policy acts according to
\begin{equation}
\label{eq:policy}
a_t \sim \pi(\cdot \mid c_t),
\qquad
\pi\ \text{fixed}.
\end{equation}

We summarize the framework-controlled runtime configuration as
\begin{equation}
\label{eq:runtime-config}
\Theta_t = (\kappa_t, \Phi_t, \iota_t),
\end{equation}
where $\kappa_t$ controls context construction, $\Phi_t$ controls semantic
acceptance and stopping, and $\iota_t$ controls diagnostic-feedback injection.
The subscript $t$ denotes the effective state-dependent behavior of each
mechanism, rather than a newly learned mechanism at every epoch.

The conceptual value of an execution trajectory is
\begin{equation}
\label{eq:runtime-objective}
J(\Theta_{0:T})
=
\mathbb{E}
\left[
\sum_{t=1}^{T}
\left(
r_t - \lambda \operatorname{cost}_t
\right)
\right],
\qquad
\lambda \ge 0,
\end{equation}
where $r_t$ and $\operatorname{cost}_t \ge 0$ denote conceptual task-utility
and runtime-cost contributions at epoch $t$, respectively. The stopping time
$T$ is determined by the acceptance and resource conditions formalized in
\S\,\ref{sec:goal}.

As new evidence becomes available, the effective runtime configuration changes
according to
\begin{equation}
\label{eq:adaptation-operator}
\Theta_{t+1}
=
\mathcal{A}\left(\Theta_t; \mathcal{M}_t\right),
\end{equation}
where $\mathcal{A}$ abstracts the state-dependent adjustments induced by the
concrete runtime mechanisms described below.

These adjustments operate within three runtime feasibility constraints:
\begin{equation}
\label{eq:C1}
\mathcal{C}_1:
\qquad
|c_t| \le B_{\mathrm{ctx}},
\qquad
\forall t,
\end{equation}
which enforces the context budget;
$\mathcal{C}_2$, which requires diagnostic feedback to be available, filtered,
and admissible before injection; and
$\mathcal{C}_3$, which requires execution to obey semantic acceptance and
configured stopping limits.

Figure~\ref{fig:runtime-adaptivity} illustrates this view. Context pressure,
semantic diagnostics, execution outcomes, goal assessments, and resource state
change the information available in $\mathcal{M}_t$ and can thereby induce a
transition from $\Theta_t$ to $\Theta_{t+1}$. The objective in
Equation~\eqref{eq:runtime-objective} is conceptual: the concrete mechanisms
implement structured online adjustments intended to steer execution toward
higher-utility feasible configurations rather than directly solving a generic
optimization program.

\subsubsection{Context Management: Adaptive Context Construction}
\label{sec:context}

Context Management realizes the context-construction component $\kappa$.
Rather than treating model context as a static concatenation of the system
prompt and complete interaction history, openJiuwen processes available
information according to current context pressure, task state, and content
structure.

Registered context processors realize $\kappa_t$ by selectively transforming,
compressing, or offloading information in $\mathcal{M}_t$ before the final
model context is assembled according to Equation~\eqref{eq:context-constructor}.
The resulting context remains subject to the budget constraint $\mathcal{C}_1$.

\paragraph{Progressive compression.}
Context is reduced progressively rather than maximally compressed at once. For
a context unit $m$,
\[
m^{(0)} \rightarrow m^{(1)} \rightarrow \cdots \rightarrow m^{(L)},
\qquad
|m^{(0)}| > |m^{(1)}| > \cdots > |m^{(L)}|,
\]
where later representations retain less detail. Concrete mechanisms include
dialogue summarization, incremental compression of large interaction histories,
and full-session compaction. Recent information is preferentially retained at
higher fidelity.

\paragraph{Structure-aware reduction and dead-loop collapse.}
Structured content such as code diffs, logs, and tables can first be reduced by
deterministic structure-aware processors before model-based summarization is
used. Repeated unproductive reasoning or tool-call patterns can likewise be
collapsed to prevent redundant history from consuming the context budget.

\paragraph{Offloading and retrieval.}
Large artifacts need not remain fully resident in active context. Their full
content can be externalized while the live context retains a compact summary or
handle. Context therefore follows a progressive information hierarchy:
\[
\text{full content}
\rightarrow
\text{structured/semantic summary}
\rightarrow
\text{compact handle}
\rightarrow
\text{on-demand retrieval}.
\]
For compressed dialogue, retrieval can proceed coarse-to-fine, first locating
relevant rounds and then recovering finer-grained source content.

Context Management can additionally coordinate with the inference layer through
session affinity and KV-cache reuse. This allows semantic context processing
and inference-state management to share session-level lifecycle information,
including during multi-agent execution.

\subsubsection{Goal Mode: Acceptance-Constrained Stopping}
\label{sec:goal}

Goal Mode realizes the acceptance and stopping component $\Phi$. It maintains
an explicit task goal across multiple execution attempts and determines whether
task-level execution should continue.

Given the framework information available at epoch $t$, the semantic completion
evaluator returns
\begin{equation}
\label{eq:phi}
\Phi_t(\mathcal{M}_t)
\in
\{
\mathsf{continue},
\mathsf{complete},
\mathsf{blocked}
\}.
\end{equation}
Semantic completion is separated from configured hard limits such as maximum
attempts, time, or resource usage. Let
$g_{\mathrm{cap}}(\mathcal{M}_t)\in\{0,1\}$ indicate whether such a limit has
been reached. The task therefore stops at
\begin{equation}
\label{eq:goal-stop}
T
=
\inf
\bigl\{
t \colon
\Phi_t(\mathcal{M}_t) \neq \mathsf{continue}
\;\lor\;
g_{\mathrm{cap}}(\mathcal{M}_t) = 1
\bigr\}.
\end{equation}
This separation distinguishes successful completion or semantic blockage from
resource exhaustion.

openJiuwen supports self-assessment, independent assessment, and hybrid
assessment strategies. Self-assessment uses the acting agent's completion
report, independent assessment evaluates the trajectory through a separate
evaluation path, and hybrid assessment combines both signals. These strategies
share the same Goal Mode interface rather than introducing different execution
engines.

Goal Mode operates within the current task. It is therefore distinct from
Self-Reflection, which acts after a trajectory has completed and affects later
tasks.

\subsubsection{LSP-Driven Passive Feedback: Closed-Loop Semantic Correction}
\label{sec:semantic}

LSP-Driven Passive Feedback realizes the diagnostic-feedback component $\iota$.
Language Server Protocol infrastructure provides mechanically derived semantic
evidence after relevant code changes, allowing newly exposed code issues to
influence subsequent model decisions.

Let $\Delta(s_t)$ denote the diagnostics produced for code state $s_t$.
Diagnostics are ranked, deduplicated, and bounded before exposure:
\begin{equation}
\label{eq:semantic-filter}
\delta_t
=
\operatorname{RankDedupLimit}
\left(
\Delta(s_t);
K_{\mathrm{file}},
K_{\mathrm{sem}}
\right),
\end{equation}
where $K_{\mathrm{file}}$ limits diagnostics contributed by one file and
$K_{\mathrm{sem}}$ limits the total diagnostic volume.

The admissible feedback is then made available to subsequent execution:
\begin{equation}
\label{eq:semantic-injection}
\mathcal{M}_{t+1}^{+}
=
\mathcal{M}_{t+1}
\cup
\iota_t(\delta_t),
\end{equation}
where $\mathcal{M}_{t+1}^{+}$ denotes the feedback-augmented information pool.
This realizes constraint $\mathcal{C}_2$: semantic feedback can affect later
decisions only after it is mechanically produced and filtered by the runtime.

The same language-server infrastructure also supports active queries such as
definition lookup, reference lookup, and call-hierarchy inspection. The
difference is control: passive diagnostics are surfaced automatically after
relevant code changes, whereas active semantic information is explicitly
requested by the agent.

The scope of this mechanism is limited to properties exposed by the underlying
analysis infrastructure, such as type errors, symbol relationships, and static
diagnostics. It does not by itself establish architectural quality,
maintainability, business-logic correctness, or end-to-end task success.

\subsubsection{Self-Reflection: Cross-Task Non-Parametric Adaptation}
\label{sec:reflection}

The mechanisms above adapt execution within one task. Self-Reflection operates
across tasks: after a trajectory has completed, it extracts reusable experience
and makes selected experience available to future executions.

For completed task $\tau_n$ with trajectory $h^{\tau_n}$, let
\begin{equation}
\label{eq:reflection-extract}
X_n
=
\Psi\left(h^{\tau_n}\right)
\end{equation}
denote the experience extracted by the configured reflection procedure. The
persistent experience store is updated through
\begin{equation}
\label{eq:reflection-update}
\mathcal{E}_{n+1}
=
\mathcal{U}\left(\mathcal{E}_n,X_n\right),
\end{equation}
where $\mathcal{U}$ may validate, deduplicate, merge, or revise extracted
experience before persistence.

For a later task $\tau_{n+1}$, let $q_{n+1}$ denote the retrieval query.
Relevant experience is retrieved as
\begin{equation}
\label{eq:reflection-retrieval}
R_{n+1}
=
\operatorname{Retrieve}
\left(
\mathcal{E}_{n+1},
q_{n+1}
\right),
\end{equation}
and becomes part of the information available to the new task:
\begin{equation}
\label{eq:reflection-pool}
\mathcal{M}_0^{+}
=
\mathcal{M}_0
\cup
R_{n+1}.
\end{equation}

Retrieval does not bypass Context Management: available experience may still be
compressed, summarized, deferred, or omitted when constructing the actual
model context.

Self-Reflection therefore changes the information available to future tasks
without updating the model policy $\pi$. It constitutes cross-task
non-parametric adaptation, complementary to the within-task runtime trajectory
illustrated in Figure~\ref{fig:runtime-adaptivity}.
\section{Experiments}
\label{sec:experiments}

\subsection{Experimental Setup}

\paragraph{Benchmarks.}
We evaluate openJiuwen on two complementary benchmarks. \textbf{SWE-bench Verified}\footnote{\url{https://www.swebench.com/}}~\cite{jimenez2024swebench} contains 500 human-validated software-engineering tasks derived from real-world GitHub issues, requiring agents to modify existing repositories and produce patches that pass the corresponding test suites. \textbf{Terminal-Bench 2.1}\footnote{\url{https://www.tbench.ai/leaderboard/terminal-bench/2.1}}~\cite{merrill2026terminalbench} contains 89 diverse tasks executed in containerized terminal environments, covering broader tool-driven and long-horizon agent execution.

\paragraph{Models and configurations.}
For \textbf{SWE-bench Verified}, we evaluate openJiuwen with \textbf{Claude Opus 4.5} using the \textbf{high} reasoning-effort setting. For the primary \textbf{Terminal-Bench 2.1} result, we use \textbf{GPT-5.6 Sol}. We additionally evaluate openJiuwen with \textbf{Fable 5}, enabling a model-matched comparison with Claude Code and Terminus 2 using the same backbone model. Within each openJiuwen configuration, the prompt, available tools, and model configuration remain fixed. SWE-bench patches are graded using the official test suites, while Terminal-Bench 2.1 tasks are evaluated using their task-specific verifiers.

\paragraph{Metrics.}
For SWE-bench Verified, we report \textbf{Pass@1}, the percentage of instances successfully resolved. For Terminal-Bench 2.1, we report \textbf{Accuracy} according to the benchmark's task-specific verifiers. For comparison with state-of-the-art systems, we use results reported on the corresponding official leaderboards.


\begin{table*}[!t]
\centering
\caption{
Comparison with leading systems on \textbf{Terminal-Bench 2.1}.
Prior results are taken from the official leaderboard.
}
\label{tab:terminal_bench_21}
\small
\renewcommand{\arraystretch}{1.18}
\setlength{\tabcolsep}{4.0pt}
\resizebox{\textwidth}{!}{%
\begin{tabular}{@{} c l l l l c c c @{}}
\toprule
\textbf{Rank} &
\textbf{Agent / Harness} &
\textbf{Agent Org.} &
\textbf{Model} &
\textbf{Model Org.} &
\textbf{Effort} &
\textbf{Date} &
\textbf{Accuracy (\%)} \\
\midrule
10 & Claude Code & Anthropic      & Sonnet 5       & Anthropic & high   & 2026-07-09 & $74.6 \pm 1.6$ \\
9  & Codex       & OpenAI         & GPT-5.6 Luna   & OpenAI    & max    & 2026-07-11 & $75.7 \pm 1.3$ \\
6  & Codex       & OpenAI         & GPT-5.6 Terra  & OpenAI    & max    & 2026-07-11 & $78.4 \pm 1.3$ \\
5  & Claude Code & Anthropic      & Opus 4.8       & Anthropic & high   & 2026-07-09 & $78.9 \pm 1.3$ \\
3  & Terminus 2  & Terminal-Bench & Fable 5        & Anthropic & high   & 2026-06-05 & $80.4 \pm 1.2$ \\
2  & Codex       & OpenAI         & GPT-5.5        & OpenAI    & xhigh  & 2026-05-01 & $83.1 \pm 1.1$ \\
1  & Claude Code & Anthropic      & Fable 5        & Anthropic & xhigh  & 2026-06-07 & $83.8 \pm 1.2$ \\
\midrule
\textbf{--} &
\textbf{openJiuwen} &
\textbf{Ours} &
\textbf{Fable 5} &
\textbf{Anthropic} &
\textbf{high} &
\textbf{2026-08-21} &
$\boldsymbol{84.04 \pm 1.12}$ \\
\textbf{--} &
\textbf{openJiuwen} &
\textbf{Ours} &
\textbf{GPT-5.6 Sol} &
\textbf{OpenAI} &
\textbf{high} &
\textbf{2026-08-21} &
$\boldsymbol{87.19 \pm 1.20}$ \\
\bottomrule
\end{tabular}%
}
\end{table*}

\begin{table*}[!t]
\centering
\caption{
Comparison with leading systems on \textbf{SWE-bench Verified}.
Bash-only mini-SWE-agent entries are excluded.
}
\label{tab:swebench_verified}
\small
\renewcommand{\arraystretch}{1.18}
\setlength{\tabcolsep}{5.0pt}
\resizebox{\textwidth}{!}{%
\begin{tabular}{@{} c l l l c c @{}}
\toprule
\textbf{Rank} &
\textbf{Agent / Harness} &
\textbf{Org.} &
\textbf{Model} &
\textbf{Date} &
\textbf{Resolved (\%)} \\
\midrule
16 & Lingxi-v1.5~\cite{yang2026lingxi}          & Lingxi & Claude 4 Sonnet          & 2025-07-20 & 74.60 \\
8  & ACoder                                    & ACoder & Multiple                 & 2025-08-19 & 76.40 \\
6  & EPAM AI/Run Developer Agent               & EPAM   & Claude 4 Sonnet          & 2025-08-04 & 76.80 \\
4  & live-SWE-agent~\cite{xia2025livesweagent} & UIUC   & Gemini 3 Pro Preview     & 2025-11-20 & 77.40 \\
3  & TRAE~\cite{traeresearchteam2025traeagent} & TRAE   & Doubao-Seed-Code         & 2025-09-28 & 78.80 \\
2  & Sonar Foundation Agent                    & Sonar  & Claude 4.5 Opus          & 2025-12-05 & 79.20 \\
1  & live-SWE-agent~\cite{xia2025livesweagent} & UIUC   & Claude 4.5 Opus & 2025-12-15 & 79.20 \\
\midrule
\textbf{--} &
\textbf{openJiuwen} &
\textbf{Ours} &
\textbf{Claude 4.5 Opus} &
\textbf{2026-08-21} &
\textbf{82.60} \\
\bottomrule
\end{tabular}%
}
\end{table*}

\subsection{Results on Terminal-Bench 2.1}

\paragraph{Results.}
As shown in Table~\ref{tab:terminal_bench_21}, openJiuwen with GPT-5.6 Sol achieves \textbf{87.19\%} accuracy on Terminal-Bench 2.1, exceeding the strongest selected result on the official leaderboard, Claude Code with Fable 5 at 83.8\%, by \textbf{3.39 percentage points}. In the model-matched Fable 5 comparison, openJiuwen also achieves \textbf{84.04\%}, slightly exceeding Claude Code at 83.8\% by \textbf{0.24 percentage points} and outperforming Terminus 2 at 80.4\%. The Fable 5 comparison reduces model capability as a confounding factor, although differences in prompts, tools, and agent implementations remain. Together, these results show that openJiuwen is competitive on broader long-horizon execution in heterogeneous terminal environments.

\begin{figure*}[!t]
    \centering
    \includegraphics[width=0.92\textwidth]{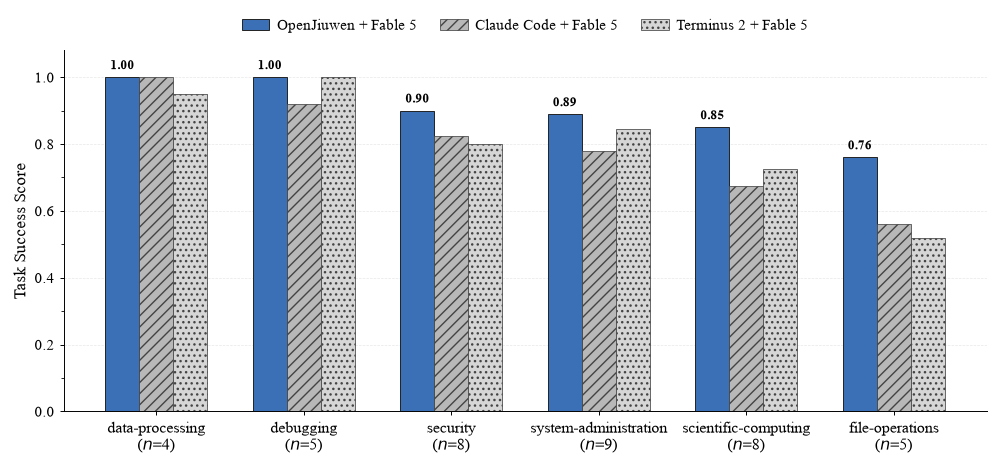}
    \caption{
    Model-matched comparison on representative Terminal-Bench 2.1 categories using Fable 5.
    openJiuwen is particularly strong on tool-intensive categories such as file operations and system administration.
    }
    \label{fig:terminal_category_strengths}
\end{figure*}

\paragraph{Category-level analysis.}
To examine where the overall result arises, Table~\ref{tab:terminal_category_breakdown} reports the complete breakdown across all 16 Terminal-Bench 2.1 task categories for both openJiuwen configurations, together with Claude Code and Terminus 2. Figure~\ref{fig:terminal_category_strengths} further highlights six representative categories in which openJiuwen with Fable 5 is particularly competitive. The Fable 5 columns provide a model-matched view that reduces differences caused by the backbone model, while the GPT-5.6 Sol column shows the category-level behavior of the primary configuration.

\begin{table*}[!t]
\centering
\caption{
Category-level task success scores on \textbf{Terminal-Bench 2.1}.
The full benchmark contains 89 tasks. Bold indicates the best score in each row (ties included).
}
\label{tab:terminal_category_breakdown}
\small
\renewcommand{\arraystretch}{1.15}
\setlength{\tabcolsep}{6pt}
\resizebox{\textwidth}{!}{%
\begin{tabular}{@{} l r c c c c @{}}
\toprule
\textbf{Task Type} &
\textbf{\# Tasks} &
\shortstack{\textbf{openJiuwen}\\\textbf{Fable 5}} &
\shortstack{\textbf{openJiuwen}\\\textbf{GPT-5.6 Sol}} &
\shortstack{\textbf{Claude Code}\\\textbf{Fable 5}} &
\shortstack{\textbf{Terminus 2}\\\textbf{Fable 5}} \\
\midrule
data-processing       & 4  & \textbf{1.00000} & \textbf{1.00000} & \textbf{1.00000} & 0.95000 \\
data-querying         & 1  & \textbf{1.00000} & \textbf{1.00000} & \textbf{1.00000} & \textbf{1.00000} \\
data-science          & 8  & 0.72500 & \textbf{0.95000} & 0.87500 & 0.87500 \\
debugging             & 5  & \textbf{1.00000} & 0.96000 & 0.92000 & \textbf{1.00000} \\
file-operations       & 5  & \textbf{0.76000} & 0.72000 & 0.56000 & 0.52000 \\
games                 & 1  & \textbf{1.00000} & \textbf{1.00000} & 0.80000 & \textbf{1.00000} \\
machine-learning      & 3  & 0.86667 & 0.86667 & \textbf{1.00000} & 0.86667 \\
mathematics           & 4  & 0.80000 & \textbf{0.95000} & 0.90000 & 0.70000 \\
model-training        & 4  & \textbf{0.80000} & 0.70000 & 0.75000 & 0.70000 \\
optimization          & 1  & \textbf{1.00000} & 0.80000 & \textbf{1.00000} & \textbf{1.00000} \\
personal-assistant    & 1  & \textbf{1.00000} & \textbf{1.00000} & \textbf{1.00000} & \textbf{1.00000} \\
scientific-computing  & 8  & \textbf{0.85000} & 0.77500 & 0.67500 & 0.72500 \\
security              & 8  & \textbf{0.90000} & 0.72500 & 0.82500 & 0.80000 \\
software-engineering  & 26 & 0.80769 & \textbf{0.90769} & 0.88462 & 0.80769 \\
system-administration & 9  & 0.88889 & \textbf{0.95556} & 0.77778 & 0.84444 \\
video-processing      & 1  & 0.20000 & 0.40000 & \textbf{0.80000} & 0.20000 \\
\bottomrule
\end{tabular}%
}
\end{table*}

The category breakdown is especially favorable on \textbf{file operations} and \textbf{system administration}. With Fable 5, openJiuwen reaches 0.76 on file operations, compared with 0.56 for Claude Code and 0.52 for Terminus 2; on system administration, it reaches 0.889, compared with 0.778 and 0.844, respectively. The GPT-5.6 Sol configuration further increases the system-administration score to 0.956 and reaches 0.908 on software engineering and 0.950 on data science. These tasks are strongly tool-dependent, requiring agents to repeatedly inspect and modify files, interact with the terminal, and verify environment state. We attribute part of openJiuwen's strength in these categories to its broad set of out-of-the-box operational tools and the unified execution interface through which these tools are exposed. This reduces the need for task-specific tool construction and allows the agent to devote more of its execution trajectory to task reasoning, action, and verification. Because the benchmark does not isolate tool availability from other harness mechanisms, this explanation should be interpreted as a plausible systems-level factor rather than a controlled causal attribution.


\subsection{Results on SWE-bench Verified}

\paragraph{Results.}
Table~\ref{tab:swebench_verified} compares openJiuwen with leading non-Bash-only systems on SWE-bench Verified. openJiuwen resolves \textbf{82.6\%} of the 500 instances, outperforming the strongest selected leaderboard results at 79.2\% by \textbf{3.4 percentage points}. Notably, the strongest selected prior systems also use Claude 4.5 Opus, suggesting that the performance difference cannot be attributed solely to the use of a stronger model. The result indicates that openJiuwen provides an effective harness for repository-level, long-horizon software-engineering tasks.

\paragraph{Performance by task duration.}
To examine robustness as the required execution horizon grows, Table~\ref{tab:swebench_duration} groups SWE-bench Verified tasks into four disjoint buckets by estimated fix duration. We focus on Claude Opus 4.5 configurations so that the comparison more directly reflects differences in the agent harness and reasoning-effort setting.

\begin{table*}[!t]
\centering
\caption{
SWE-bench Verified success rates grouped by estimated fix duration.
Bold indicates the best score in each row (ties included).
}
\label{tab:swebench_duration}
\footnotesize
\renewcommand{\arraystretch}{1.2}
\setlength{\tabcolsep}{5.0pt}
\resizebox{\textwidth}{!}{%
\begin{tabular}{@{} l r c c c c c @{}}
\toprule
\textbf{Fix Duration} &
\textbf{\# Tasks} &
\shortstack{\textbf{Opus 4.5 (high)}\\\scriptsize mini-swe-agent} &
\shortstack{\textbf{Opus 4.5 (med.)}\\\scriptsize mini-swe-agent} &
\shortstack{\textbf{Opus 4.5 (med.)}\\\scriptsize live-SWE-agent} &
\shortstack{\textbf{Opus 4.5 (med.)}\\\scriptsize Sonar Foundation} &
\shortstack{\textbf{Opus 4.5 (high)}\\\scriptsize \textbf{openJiuwen}} \\
\midrule
$<15$ min fix
& 194
& 0.89175
& 0.84536
& 0.88660
& 0.88144
& \textbf{0.91753} \\

15 min--1 hour
& 261
& 0.74713
& 0.72414
& 0.76628
& 0.77778
& \textbf{0.81226} \\

1--4 hours
& 42
& 0.35714
& 0.42857
& \textbf{0.54762}
& 0.50000
& 0.52381 \\

$>4$ hours
& 3
& \textbf{0.33333}
& \textbf{0.33333}
& \textbf{0.33333}
& \textbf{0.33333}
& \textbf{0.33333} \\
\bottomrule
\end{tabular}%
}
\end{table*}

The duration breakdown shows that openJiuwen remains competitive as the execution horizon grows. It achieves the best result on both the $<15$ minute bucket (91.75\%) and the substantially larger 15-minute--1-hour bucket (81.23\%). On the 42 tasks in the 1--4 hour bucket, openJiuwen reaches 52.38\%, substantially above mini-swe-agent with the same Opus 4.5 high setting (35.71\%) and also above mini-swe-agent with medium reasoning effort (42.86\%); it is close to the strongest result from live-SWE-agent at 54.76\%. The $>4$ hour bucket contains only three tasks, so we do not draw a strong conclusion from that row alone.

An interesting pattern appears within mini-swe-agent: Opus 4.5 with \emph{high} reasoning effort underperforms its \emph{medium} setting on the 1--4 hour tasks (35.71\% vs.\ 42.86\%). A plausible explanation is a context-budget effect. Higher reasoning effort can spend more tokens on deliberation, and over long trajectories this can increase pressure on the finite context window, leaving less effective capacity for repository state, tool feedback, and execution history. In contrast, openJiuwen uses Opus 4.5 with high reasoning effort yet reaches 52.38\% on the same duration bucket. This result is consistent with the role of openJiuwen's context-management mechanism, which selectively retains and exposes task-relevant information as execution proceeds, mitigating context pressure while preserving the benefit of deeper reasoning. Since the benchmark does not directly isolate reasoning-token usage or context overflow, we present this as a systems-level interpretation rather than a controlled causal claim.

\paragraph{Overall comparison.}
Across both benchmarks, openJiuwen exceeds the strongest selected leaderboard result: by \textbf{3.39 percentage points} on Terminal-Bench 2.1 and by \textbf{3.4 percentage points} on SWE-bench Verified. The two benchmarks stress substantially different execution settings---repository-grounded issue resolution and general terminal-based task execution---yet openJiuwen achieves strong performance in both. This consistency supports the design goal of openJiuwen as a reusable harness for coordinating complex agent execution across heterogeneous long-horizon workloads. Because leaderboard systems may differ in models, prompts, tools, and agent implementations, these results should be interpreted as system-level comparisons rather than controlled estimates of the contribution of the harness alone.
\section{Limitations}
\label{sec:limitations}

While openJiuwen provides a general harness for structurally composable and
runtime-adaptive coding agents, several directions remain for further
development. First, the current design treats runtime Self-Reflection and
offline evolution as complementary stages; tighter integration between them
could enable more systematic reuse of execution experience. Second, Goal Mode
currently focuses on task-level objectives, and extending it to hierarchical
goals may further improve control over complex, multi-stage tasks. Third,
future work can further automate the selection and coordination of Context
Management strategies under different runtime conditions. Finally, our current
evaluation focuses on SWE-bench Verified and Terminal-Bench 2.1. Broader
controlled studies across additional benchmarks, models, and configurations,
together with more detailed ablations, would further characterize the
generality and contributions of individual harness mechanisms.
\section{Conclusion}
\label{sec:conclusion}

We presented \textbf{openJiuwen}, an \textbf{open-source coding-agent harness} that addresses two complementary challenges: \textbf{Structural Composability} and \textbf{Runtime Adaptivity}. For developers, openJiuwen provides a shared \textbf{Inner Loop/Outer Loop} execution substrate, \textbf{Rail}-based capability composition, and reusable execution semantics across single agents, delegated sub-agents, and multi-agent coordination, reducing the engineering effort required to construct and extend sophisticated agent harnesses. For complex coding tasks, openJiuwen adapts framework-controlled runtime decisions through \textbf{Goal Mode}, \textbf{LSP-Driven Passive Feedback}, \textbf{Context Management}, and \textbf{Self-Reflection}, allowing evolving execution evidence to continuously guide subsequent behavior. Evaluations on \textbf{SWE-bench Verified} and \textbf{Terminal-Bench 2.1} demonstrate strong performance across repository-level software engineering and broader terminal-based tasks. Together, these results show that a composable harness architecture can support both efficient agent development and effective execution on complex coding tasks.


\bibliography{main}
\bibliographystyle{plainnat}

\appendix

\section{Author List}
\label{app:extra}

\textbf{Core Contributors.} Tao Yu, Xinyu Zhang, Qianqian Chen, Xiaoneng Xiang, Chia Kwangyang, Xingchen Huang, Ran Chen, Yangkai Ding, Zheng Wang, Yeo Boon Hong, Bingzheng Gan, Enrui Hu, Shuo Cheng, Deyang Li, Ruifeng Shi, Hongbo Wang, Qi Ye, Xuefeng Jin, Zhangchun Zhao

\end{document}